\documentclass[12pt,a4paper]{article}
\usepackage{graphicx}
\usepackage{subfigure}
\usepackage{makecell,rotating,multirow,diagbox}
\usepackage{cavw}
\usepackage[noblocks]{authblk}
\begin{document}
\title{Improved Face Detection and Alignment using Cascade Deep Convolutional Network}
\author[$\dagger$]{Weilin Cong}
\author[$\dagger$]{Sanyuan Zhao}
\author[$\ddag$]{Hui Tian}
\author[$\dagger$]{Jianbing Shen}
\affil[$\dagger$]{Beijing Key Laboratory of Intelligent Information Technology, School of Computer Science,Beijing Institute of Technology, Beijing 100081, P.R.China}
\affil[$\ddag$]{China Mobile Research Institute, Xuanwu Men West Street, Beijing}
\maketitle
\vspace{-30 pt}
\begin{abstract}
Real-world face detection and alignment demand an advanced discriminative model to address challenges by pose, lighting and expression. Recent studies have utilized the relation between face detection and alignment to make models computationally efficiency, but they ignore the connection between each cascade CNNs. In this paper, we combine detection, calibration and alignment in each Cascade CNN and propose an HEM method for \emph{End-to-End} cascade network training, which give computers more space to automatic adjust weight parameter and accelerate convergence. Experiments on FDDB and AFLW demonstrate considerable improvement on accuracy and speed.
\end{abstract}
\section{Introduction}
Face detection and alignment have play an important role in many face-based applications, challenges mainly come from various poses, uncontrollable illuminations and exaggerated expressions. Inspired by the remarkable performance of deep learning, some of the convolution based feature extraction methods have been proposed in recent years. In face detection area, Li \emph{et al.}\cite{Cascade CNN} use cascade CNNs for face detection. An extra calibration stage is added after each detection stage which cost extra computing expense on bounding box calibration. Yang \emph{et al.}\cite{From facial parts responses to face detection} extracte the features of hair, eyes, nose, mouth and neck regions with five separate CNNs, combine the result of five CNNs and use the position information to improve face detection result. Due to it's complex structure, this approach is time costly in practice. Zhang \emph{et al.}\cite{mtcnn} proposed a framework that joint detection and alignmnet in single CNN, however it is not an end-to-end structure and the model is very complicated to train. Face alignment also attracts extensive interests. Sun \emph{et al.}\cite{DCNNs Cascade points} use three-level CNNs extract global and local features to estimate the positions of facial key points. Zhou \emph{et al.}\cite{Extensive Cascade points} use coarse-to-fine four-level cascade architecture extend 5 points detection to 68 key points detection in real-time detection. 

Since detection always followed by alignment, we adopt a structure to calculate detection, calibration with alignmnet simultaneously and propose an End-to-End strategy to unify the Cascade structure. Hard Example Proposal Network is used to generate high-quality proposals. 

This model is tiny but fast. The summary size of models is 2MB, much smaller than well known ImageNet Challenge models like VGG16(is about 510MB) and AlexNet(is about 250MB). For 800 $\times$ 600 images, our detector can detect faces and key points 90 fps on Titan X, faster than most state-of-the-art models\footnote{FRCN can detect object 7 fps, SSD detect object 59 fps}. The contributions of this paper are summarized as follows: (1)We propose an Hard Sample Proposal Structure to automatic generate higher quality training samples simultaneously for each cascade CNN. (2)We propose an End-to-End architecture and combine detection, calibration with alignment to make system more robust and accurate. (3)Extensive experiments on benchmarks show impressive improvement than the state-of-the-art face detection and alignmnent tasks.

\begin{figure}[htb]
\setlength{\abovecaptionskip}{-0.2cm}
\includegraphics[width=1\textwidth]{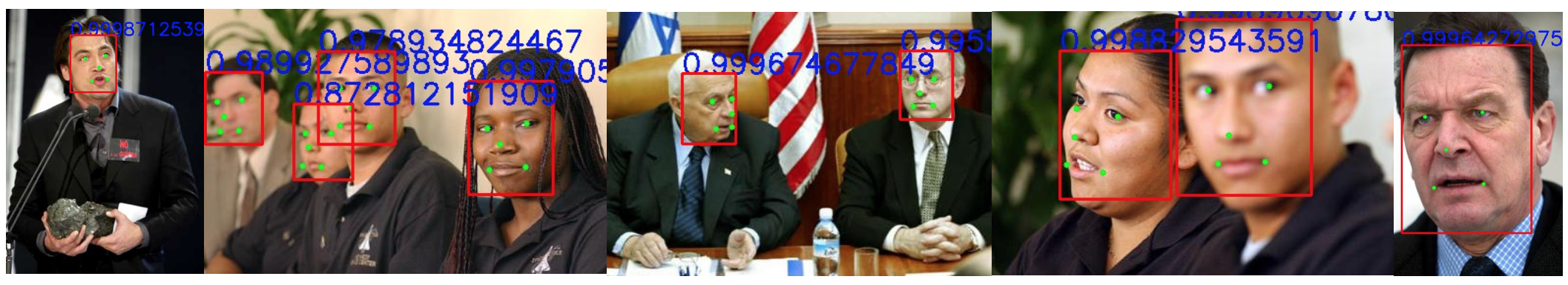}
\caption{Selected examples of face detection and key points detection on FDDB test set. A score threshold of [0.6,0.6,0.7] is used for 12net, 24net and 48net to display these images. The running time for obtaining these results is 15ms per image in average.}\label{fig:fddb test}
\end{figure}

\section{Approach}
\subsection{Overall framework}
\begin{figure}[htb]
\setlength{\abovecaptionskip}{-0.2cm}
\includegraphics[width=1\textwidth]{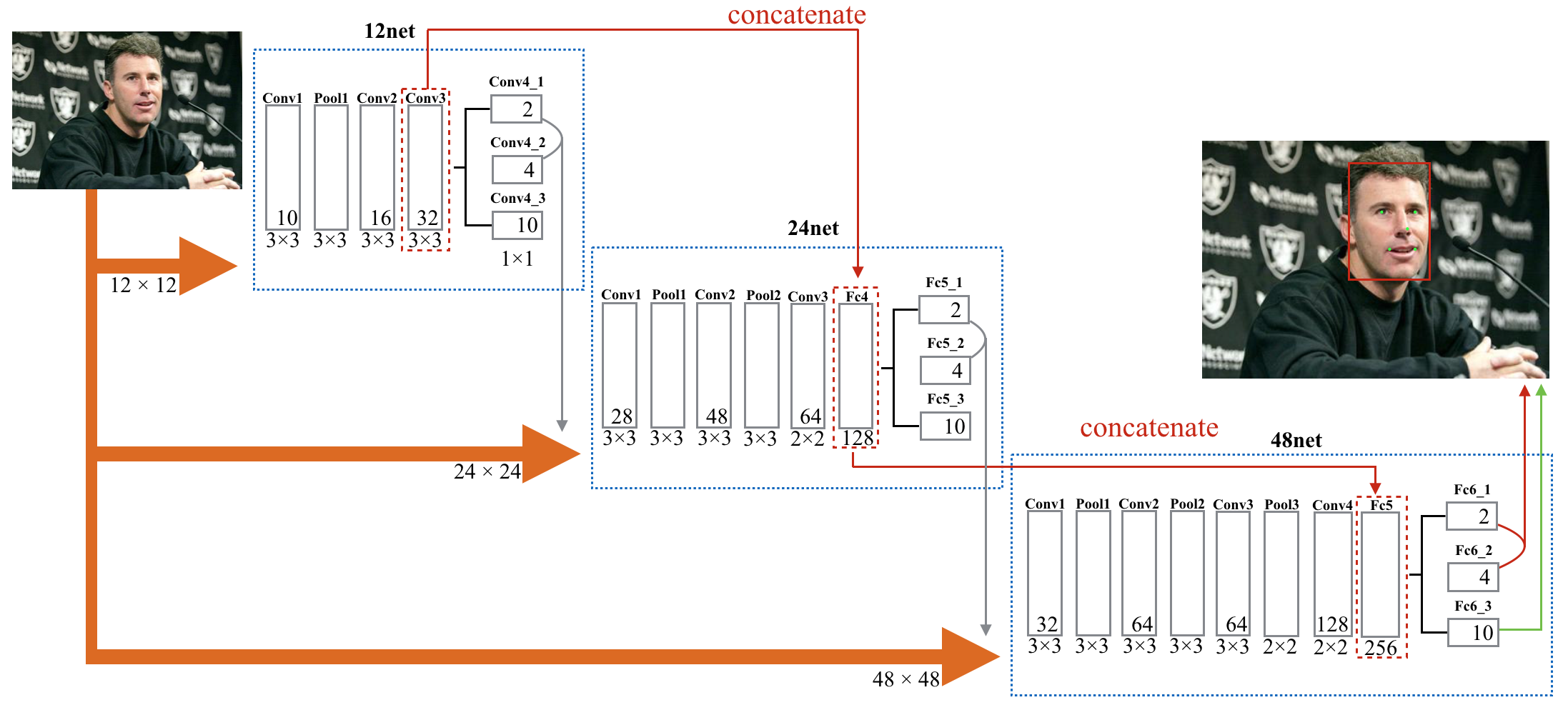}
\caption{Pipeline of our detector: 12net produce multi candidate windows, 24net refine these candidate windows and 48net produces final bounding box and facial landmarks position. Each single CNN produce three different outputs: face vs non-face classification result, candidate windows calibration result and key points detection result. An End-to-End strategy is adopted between 12net and 24net, 24net and 48net. }\label{fig:pipline}
\end{figure}

The overall pipeline of our framework is shown in Figure \ref{fig:pipline}. Given a test image, 12net use fully convolution method look through the image only one time to remove most of the negative candidates and calibrate positive candidates. Local-NMS is applied to eliminate candidate windows with high overlap ratio. Then remaining candidate windows are cropped and scaled to $24 \times 24$ for 24net to remove more negative candidates and calibrate positive ones, Local-NMS is also applied to further reduce the number of candidates. Finally, 24net's outputs are cropped and scaled to $48 \times 48$ for 48net to estimate the location of the last candidate windows and facial key points. Global-NMS will eliminate the candidate windows with higher proportion IoU and output the final results.

\subsection{Loss Function}
We consider three parts for loss functions: face vs non-face classification loss, candidate windows regression loss and facial key points detection loss. We minimize a muiti task loss defined as:
\begin{eqnarray}
L({cls}_i,{reg}_i,{pts}_i)  = \alpha \sum L_{cls}({cls}_i,{cls}_i^*)  +  \beta \sum L_{reg}({reg}_i,{reg}_i^*)  +  \gamma \sum L_{pts}({pts}_i,{pts}_i^*) )
\end{eqnarray}
Where $\alpha$, $\beta$, $\gamma$ are used to balance different task loss. In 12net and 24net we keep $\alpha$ = 1,$\beta$ = 0.5,$\gamma$ = 0.5 while in 48net $\alpha$ = 1,$\beta$ = 0.5,$\gamma$ = 1.

The face vs non-face classification task can be regarded as binary classification problem. For each sample $x_i$, we use logistic regression loss after softmax layer:
\begin{eqnarray}
L_{cls}({y}_i,{y}_i^*) = -(y_i^* log(y_i)+(1-y_i^*) log(1-y_i))
\end{eqnarray}
Where $y_i$ is the probability of being a face, $y_i^* \in\{0,1\}$ denote ground truth labels.

For each candidate window, we predict the offset between it and the nearest ground truth. For each sample $x_i$, we use SmoothL1 loss:
\begin{eqnarray}
L_{reg}({y}_i,{y}_i^*)=\left\{
\begin{array}{rcl}
0.5x^2              &  & \textrm{if $|x| < 1$} \\
\left|x\right|-0.5  &  & \textrm{else}
\end{array} \right. 
\end{eqnarray}
Where $y_i$ is the predicted offset, $y_i^* = \{x,y,w,h\}$  is the ground truth offset.

Similar to the candidate windows regression task, facial key points detection is seen as a regression problem and we minimize the SmoothL1 loss:
\begin{eqnarray}
L_{pts}({y}_i,{y}_i^*)=\left\{
\begin{array}{rcl}
0.5x^2              &  & \textrm{if $|x| < 1$} \\
\left|x\right|-0.5  &  & \textrm{else}
\end{array} \right. 
\end{eqnarray}
Where $y_i$ is the facial key points predicted by the network and $y_i^* = \{x_1,y_1,x_2,y_2,...,x_5,y_5\}$ is the ground truth five facial landmarks positive.
\subsection{Multi-task Training}
It has been proved by Li \emph{et al.}\cite{Cascade CNN} that adding a calibration function CNN after detection function CNN can improve candidate window localization effectiveness and significantly reduce the number of candidate windows. However, calibration CNNs requires extra computational expense and cause cascade structure too complex to handle. In order to reduce the amount of calculation and simplified the structure of the cascade CNNs, we joint detection with calibration similar to Faster-RCNN.

In addition, Chen \emph{et al.}\cite{joint cascade cnn} use a face alignment model detect the face key points and combine face alignment with face detection, the result has proved that aligned face shapes provide better performance to face detection. Since seperating key points location and face detection will cause more computational expense, we combine key points detection to provide the cascade CNNs with information from different aspect, and this approach lead the model a better performance. The result can be shown in Figure \ref{fig:joint training}.

\label{mini batch}In order to train this multi-loss CNN, for each iteration batch we have to select different training data for different loss and only backpropagrate NN with specific data type and restrain others. There should be four different kinds of training data(including \emph{Positives}, \emph{Negatives}, \emph{Part faces} and \emph{Key Points Detection} data) for three tasks. Negatives and positives are used for face vs non-face classification task, positives and part faces are used for candidate windows regression and landmark faces are used for facial landmarks detection. In this paper, Positives, Negatives, Part Faces data are generated on \emph{Wider Face}\cite{wider face} and key points detection data is generated on \emph{CelebA}\cite{celeba}.

 We can either a) have a large batch size( = 256) and randomly select different kinds of data for this batch or b) have several iteration mini-batchs( = 64) and for each mini-batch have same kind of training data. According to Table \ref{batch size compare}, training with mini-batchs can have less calculations, faster training speed and similar accuracy to large batch option. So we choose mini batch in our training approach.
\begin{table}
\centering
\caption{Compare Large-batch training and Mini-batch training}
\begin{tabular}{|c|c|c|}
\hline
\diagbox {Network}{Evaluation} & 1000 Times Calculate(large/mini batch) & Accuarcy(large/mini batch) \\
\hline
12net & 0.117s / 0.103s & 94.4\% / 94.4\% \\
\hline
24net & 1.521s / 1.510s & 95.4\% / 95.1\% \\
\hline
48net & 4.701s / 4.490s & 95.2\% / 95.2\% \\
\hline
\end{tabular}
\label{batch size compare}
\end{table}
\subsection{Hard Example Proposal Network}
Deep learning process has achieved significant advances because of big datasets, but still includes many heuristics and hyperparameters that are difficult to select. On the other hand, big datasets always contain an large number of easy examples and a small number of hard examples, using too many easy examples will not improve much to the result of our CNN models. Therefore, reference to object detection hem methods\cite{OHEM,faster rcnn}, we design a \emph{Hard Example Proposal Network} for cascade, which generate the hard training samples simultaneously for three seperate CNNs.

\begin{figure}[htb]
\setlength{\abovecaptionskip}{-0.2cm}
\includegraphics[width=1\textwidth]{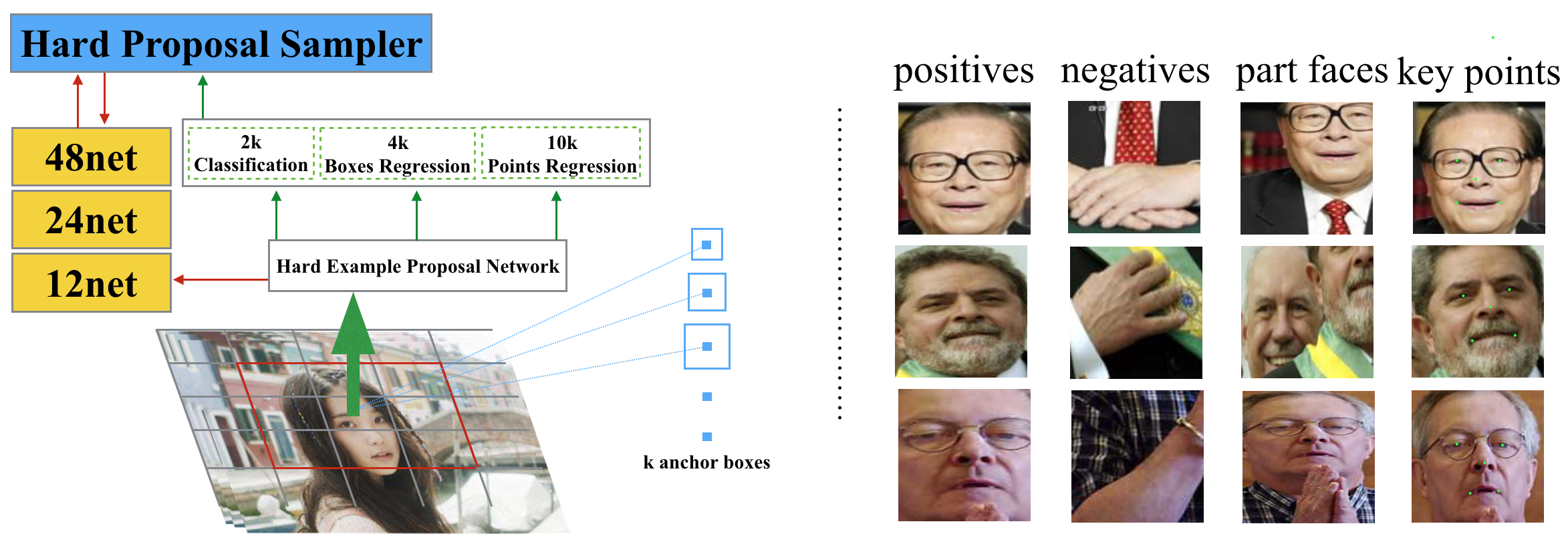}
\caption{Hard Example Proposal Network generate hard proposals and labels during training process. According to loss, Hard Proposal Sampler choose only top 70\% of training data backward propagrate.}\label{fig:rpn}
\end{figure}

As shown in Figure \ref{fig:rpn}, Hard Example Proposal Network regular generate rectangular object proposals, each of them have an object class according to the IoU. The proposals which have IoU ratio less than 0.3 to all ground truth is classified as Negatives, the proposals which have IoU ratio above 0.7 to a ground truth is classified as Positives, the proposals with IoU between 0.3 and 0.7 is recognized as Part faces. Hard Example Proposal Network also crop the ground truth and label five landmarks' position as Landmark faces.

In Hard Example Proposal Network, three different loss layers respectively compute loss for their training data, sort them from large to small based on loss. The network will automatically select the top N(\emph{batch size} = N) as hard examples and set the loss of all non-hard examples to 0. If hard example is not enough, then randomly transform hard examples to fill in batches. Experiments shows that this stategy lead training converge faster and a better performance. It is proved to be surprisingly effective in accelerating convergence and reducing training loss Figure \ref{fig:ohem}.
\subsection{End To End Training.}
Compared to other cascade convolutional neural network structure\cite{joint cascade cnn, mtcnn,DCNNs Cascade points,Cascade CNN,Extensive Cascade points}, we propose an \emph{End-to-End} framework to train the cascade CNNs once a time. End-to-End training is to give the network model a raw input to obtain a final output, it can reduce the manual preprocessing and subsequent processing, the model can automatically adjust itself according to the data as much as possible, thereby increasing the overall fit of the model.

In ordinary cascade CNNs architecture\cite{Cascade CNN}, independent cascade CNN do not share weight and bias with other CNNs, it's impossible to update multi cascade CNNs during one forward propagate and backward propagate. In order to get rid of this limitation, as shown in Figure \ref{fig:pipline}, we build a bridge between 12net and 24net, 24net and 48net. With this concatenate structure, the weight matrix and bias matrix is concatenated to higher cascade level CNN, the higher cascade level CNN is supplemented by the information from lower cascade level CNN which enable the detector to learn multi resolution information and offer a chance to backward propagate from 48net to 12net.

In this paper, an \emph{Alternating Training} method is proposed to achieve End-to-End training. At the very beginning, 12net, 24net and 48net are trained independently, they modify their convolution layers in different ways. Then the convolution layer's weigh and bias parameters are extracted and duplicated to an End-to-End Cascade structure, and fine-tune them on same dataset with Hard Example Proposal Network. At this time, three seperate CNNs will share same weight, become more unified and robust.
\section{Experiments}
\begin{figure}[htb]
\setlength{\abovecaptionskip}{-0.1cm}
\centering
\subfigure[Joint training]{
    \label{fig:joint training}
    \begin{minipage}{4cm}
    \centering
    \includegraphics[width= 4cm]{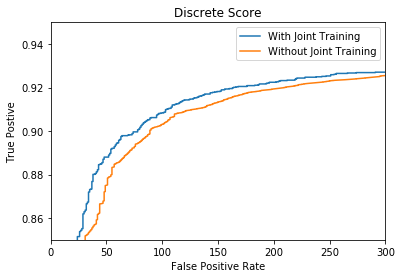}
    \end{minipage}
    }
\subfigure[OHEM training]{
    \label{fig:ohem}
    \begin{minipage}{4cm}
    \centering
    \includegraphics[width= 4cm]{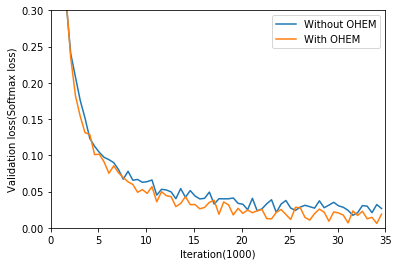}
    \end{minipage}
    }
\subfigure[End-to-End training]{
    \label{fig:end2end}
    \begin{minipage}{4cm}
    \centering
    \includegraphics[width= 4cm]{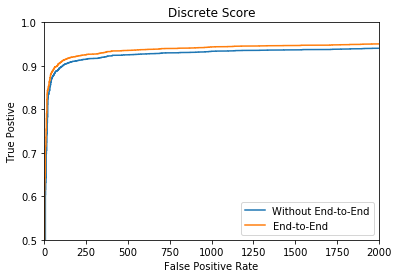}
    \end{minipage}
    }
\caption{The figure shows the contribution of joint training strategy, online hard example mining strategy and end-to-end training strategy on FDDB\cite{fddb} and AFLW\cite{aflw} Benchmark}
\end{figure}
\subsection{Effectiveness of Joint Training}
To evaluate the contribution of joint training strategy on FDDB, we fix the 12net, 24net and modity the 48net to joint and non-joint versions. Figure \ref{fig:joint training} suggests that jointing face alignment with face detection task is beneficial for face detection.
\subsection{Effectiveness of OHEM in Hard Example Proposal Network}
To evaluate the effectiveness of OHEM(online hard example mining) training strategy in Hard Example Proposal Network, we trained two 48net version(OHEM training version and Without-OHEM training version) on same dataset. Both of them have same learning rate(\emph{lr = 0.01}), batch size(\emph{batch size = 64}) and trained on Wider Face Training dataset\cite{wider face}. Figure \ref{fig:ohem} represent the loss curves of two different training ways. According to the loss curves, OHEM can accelerate convergence and lead to a lower loss value. It is easy to draw a conclusion that Hard Example Proposal Network, especially OHEM, is beneficial to performance improvement.
\subsection{Effectiveness of End-to-End}
To evaluate the effectiveness of End-to-End training strategy, the same 12net, 24net and 48net initialize models are used to generate End-to-End and non End-to-End training versions. Figure \ref{fig:end2end} represent the face detection result of two different training strategies on FDDB. End-to-End structure can lead to a more unified and robust system. The ROC curve suggests that End-to-End is beneficial to cascade CNNs structure.
\begin{figure}[htb]
\setlength{\abovecaptionskip}{-0.2cm}
\centering
\subfigure[FDDB result]{
    \label{fig:FDDB}
    \begin{minipage}{7cm}
    \centering
    \includegraphics[width= 7cm]{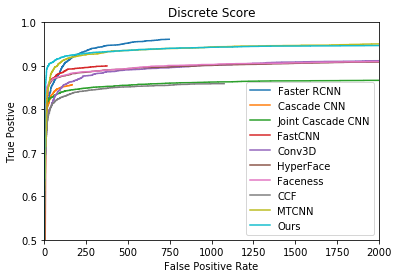}
    \end{minipage}
    }
\subfigure[AFLW result]{
    \label{fig:aflw}
    \begin{minipage}{7cm}
    \centering
    \includegraphics[width= 7cm]{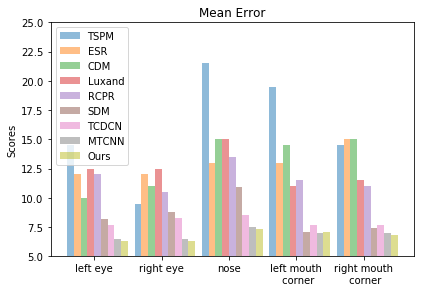}
    \end{minipage}
    }
\caption{The figure shows the comparation of our model with other state-of-art methods \cite{faster rcnn, Cascade CNN,mtcnn,joint cascade cnn,fast rcnn,conv3d,ccf,hyper face,faceness} on FDDB and \cite{TSPM,ESR,CDM,Luxand face SDK,RCPR,SDM,TCDCN} on AFLW.}
\end{figure}
\subsection{Evaluation on FDDB}
FDDB is a face detection dataset and benchmark, it contains 5,171 annotated faces in 2,845 images. We compare our model with other state-of-the-art methods\cite{faster rcnn, Cascade CNN,mtcnn,joint cascade cnn,fast rcnn,conv3d,ccf,hyper face,faceness} on FDDB, the ROC result is shown in Figure \ref{fig:FDDB}.
\subsection{Evaluation on AFLW}
 AFLW is a facial key points detection benchmark which contains 24,386 faces with facial landmarks. We compare our model with other state-of-the-art methods\cite{TSPM,ESR,CDM,Luxand face SDK,RCPR,SDM,TCDCN} on AFLW, the mean error result is shown in Figure \ref{fig:aflw}.
 
\section{Conclusion}
In this paper, we have proposed an End-to-End Multi-Output Cascade CNNs and introduce details on how to achieve. Experiments demonstrate that our framework can cope with the complex environment in reality and perform well on several challenging benchmarks(including FDDB\cite{fddb} for face detection, and AFLW\cite{aflw} for face alignment). In the future, we will make more efforts on the correlation between face detection and object detection or other aspects to further improve the performance. The further improvement will be published in the future.

\end{document}